# Arabic Chatbot Technologies in Education: An Overview

Hicham BOURHIL [1][0009-0007-7053-6918]
and Yacine EL YOUNOUSSI [1][0000-0003-4081-8248]

[1] AbdelMalek Essaadi University, 93030 Tetouan, Morocco

hicham.bourhil@etu.uae.ac.ma

**ABSTRACT.** The recent advancements in Artificial Intelligence (AI) in general, and in Natural Language Processing (NLP) in particular, and some of its applications such as chatbots, have led to their implementation in different domains like education, healthcare, tourism, and customer service. Since the COVID-19 pandemic, there has been an increasing interest in these digital technologies to allow and enhance remote access. In education, e-learning systems have been massively adopted worldwide. The emergence of Large Language Models (LLM) such as BERT (Bidirectional Encoder Representations from Transformers) and GPT (Generative Pre-trained Transformers) made chatbots even more popular. In this study, we present a survey on existing Arabic chatbots in education and their different characteristics such as the adopted approaches, language variety, and metrics used to measure their performance. We were able to identified some research gaps when we discovered that, despite the success of chatbots in other languages such as English, only a few educational Arabic chatbots used modern techniques. Finally, we discuss future directions of research in this field.

**KEYWORDS.** Artificial Intelligence, Educational Chatbots, Arabic language, Natural Language Processing.

## 1. Introduction

A Chatbot, also called a conversational agent or dialogue system, is an application of Artificial Intelligence that uses Natural Language processing techniques to automatically communicate and interact with users using text or speech. It has gained much interest from research and industry in the last few years.

The massive adoption of e-learning technologies since the COVID-19 pandemic has led to important technological innovations. Educational chatbots can contribute to the advancement of remote learning experience. This may save significant time and costs for institutional employees and provide more efficient and personalized learning for students. [1]

While conversational systems using the English language have achieved a good level of maturity, there is still a considerable gap in Arabic systems [2]. However, the Arabic language is the fourth most used language on the internet and is spoken by more than 400 million people around the world [3].

There are three main categories of Arabic [4]:

• Classical Arabic (CA) is an old form of the language, used in the Quran (the Islamic sacred book), as well as in the poetry and literature
• Modern Standard Arabic (MSA) is the modern variety used in today's formal communications, such as the news

• Dialect Arabic (DA) is the colloquial form of Arabic, which varies according to the region. Examples include Egyptian, Levantine, Gulf, Iraqi, and Maghrebi.

According to [5], there are 3 types of chatbots:

• Rule-based chatbots use a predefined set of rules to convert user input to convenient output. This approach allows only a fixed set of responses.
• Retrieval-based chatbots also provide predefined responses but use heuristics to retrieve the best possible response from a corpus. This approach could be based on AIML (Artificial Intelligence Markup Language) files which are based on the XML format to represent knowledge.
• Generation-based chatbots can generate new responses using advanced techniques of AI such as machine learning and deep learning. This approach represents a state-of-the-art solution [6].

It is important to note that some studies use pre-built Frameworks such as Google DialogFlow [7] and RASA [8].

## 2. Research problem and aim

According to a scoping review on Arabic chatbots [5], only 13 Arabic chatbots were identified without specifying the application domain, of which only one was based on generative AI. The rest were retrieval-based, meaning they are relatively less advanced since they do not generate new responses.

Another review [9], which was done in 2022, identified 50 studies about Arabic chatbots, of which 15 were used in education. However, some of these studies represent one common bot, which means that their number was less than 15 unique systems.

These two reviews were made in 2022, which means that a new search is needed for related studies from 2022 up to the moment of this research (August 2024).

Therefore, our objective is to address this gap by identifying the existing educational Arabic chatbots in literature. Furthermore, we aim to give an overview of their characteristics such as the adopted approaches, the supported language varieties (Classical Arabic, MSA, or Dialects), and the metrics used (Human-based or Automatic). To the best of our knowledge, this is the first survey that focuses on Educational Arabic chatbots.

## 3. Educational Arabic Chatbots

We identified 8 unique educational Arabic chatbots from the two reviews mentioned above ([5] and [9]). Additionally, we identified two more recent bots ([4] and [8]). Thus the total number of systems found is 10.

We noticed that almost all chatbots use Modern Standard Arabic (MSA), while only one [10] supports Classical Arabic, and another one [8] supports Saudi Dialect.

Furthermore, 7 of the 10 bots are retrieval-based (using the Pattern Matching technique). Two chatbots were developed using existing frameworks. Only one system [4] is based on generative AI, which is the most advanced technique.

Another important aspect found in this study is that most of the metrics used are based on human feedback to measure user satisfaction. This method may introduce significant bias due to its subjective nature. Only the two most recent chatbots used Automated-based metrics (Accuracy, F1-score, precision, and BLUE).

The following Table 1 shows a summary of all the bots found.

13

| Chatbot Name | Reference | Country | Language | Approach | Metrics |
|---|---|---|---|---|---|
| Abdullah | Alobaidi (2013 [10], 2015 [11]) | UK | CA, MSA | Retrieval-based | User Satisfaction |
| ArabChat | Hijjawi (2014 [12], 2015 [13], 2016 [14]) | Jordan | MSA | Retrieval-based | User Satisfaction |
| LANA | Aljameel (2017 [15], 2019 [16]) | Saudi Arabia | MSA | Retrieval-based | User Satisfaction |
| LABEEB | Almurtadha (2019) [17] | Saudi Arabia | MSA | Retrieval-based | N/A |
| SALAM | ElGibreen (2020) [18] | Saudi Arabia | MSA | Retrieval-based | User Satisfaction |
| Jooka | El Hefny (2021) [7] | Egypt | MSA | Framework | User Satisfaction |
| SIAAA-C | Sweidan (2021) [19] | Jordan | MSA | Retrieval-based | User Satisfaction |
| SEG-COVID | Sweidan (2021) [20] | Jordan | MSA | Retrieval-based | User Satisfaction |
| N/A | Alqahtani (2023) [8] | Saudi Arabia | Saudi Dialect | Framework | Accuracy, F1-score, and precision |
| N/A | Alazzam (2023) [4] | UAE | MSA | Generation-based | Accuracy, F1-score and BLUE |

Table 1. Educational Arabic chatbots and their characteristics.

## 4. Conclusion

Chatbots have become very popular in education and other fields due to factors such as the COVID-19 pandemic and the emergence of pre-trained models such as GPT and BERT. This paper showed that educational Arabic chatbots are still scarce and mostly immature. Researchers can address this gap by taking advantage of the most advanced techniques in AI and NLP such as Deep Learning, by using automated and standard metrics to set a unified benchmark. Moreover, Classical Arabic and Dialects are rarely supported in Arabic chatbots due to the paucity of Corpora, hence the need for contributions to building new large Arabic datasets.

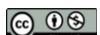